\documentclass[a4paper, 10pt, conference]{ieeeconf}      

\IEEEoverridecommandlockouts                              

\usepackage{graphics} 
    \DeclareGraphicsExtensions{.jpg,.pdf,.mps,.png} 
    \graphicspath{{fig/} {./}} 
\usepackage{epsfig} 
\usepackage{amsmath} 
\usepackage{amssymb}  
\usepackage{setspace} 
\usepackage{bm} 
\usepackage[flushmargin,hang]{footmisc} 
\usepackage{flushend} 

\newcommand{\fig}[1]{Fig.~\ref{#1}}

\newcommand{\tab}[1]{Table~\ref{#1}}

\newcommand{\eq}[1]{(\ref{#1})}

\def\epsgaiji#1{\leavevmode\kern-0.025zw\raise-.37zh\hbox{%
  \epsfile{file=#1,width=1.05zw}}\kern-0.025zw}
\newcommand{\MARU}[1]{{\ooalign{\hfil#1\/\hfil\crcr\raise.167ex\hbox{\mathhexbox20D}}}}

\usepackage{array}

\usepackage{siunitx} 
\usepackage{gensymb} 
\usepackage{multirow} 
\usepackage{caption}
\usepackage{subcaption}
\usepackage{colortbl} 
\usepackage{xcolor} 

\usepackage{pgfplots}
\pgfplotsset{compat=newest}
\usetikzlibrary{plotmarks}
\usetikzlibrary{arrows.meta}
\usepgfplotslibrary{patchplots}
\usepackage{xcolor}

\usepackage{multirow}    
\usepackage{tabularx}    
\usepackage{stfloats}    

\usepackage[export]{adjustbox}

\usepackage{grffile}
\pgfplotsset{plot coordinates/math parser=false}
\newlength\fwidth
\newlength\fheight

\usepackage{cite}

\title{\LARGE \bf
Towards the Automation in the Space Station: Feasibility Study \\and Ground Tests of a Multi-Limbed Intra-Vehicular Robot}
\author{Seiko Piotr Yamaguchi$^{*1}$, Kentaro Uno$^{*2}$, Yasumaru Fujii$^{3}$,\\ Masazumi Imai$^{2}$, Kazuki Takada$^{2}$, Taku Okawara$^{2}$, Kazuya Yoshida$^{2}$
\thanks{$^{1}$S. P. Yamaguchi is with Japan Aerospace Exploration Agency (JAXA), Tsukuba, Japan, and Space Robotics Lab (SRL) in the Department of Aerospace Engineering, Graduate School of Engineering, Tohoku University, Sendai, Japan. (E-mail: {\tt\small yamaguchi.seiko@jaxa.jp}) }%
\thanks{$^{2}$K. Uno, M. Imai, K. Takada, T. Okawara, and K. Yoshida are with the Space Robotics Lab. (SRL) in the Department of Aerospace Engineering, Graduate School of Engineering, Tohoku University, Sendai, Japan. (E-mail: {\tt\small unoken@tohoku.ac.jp})}%
\thanks{$^{3}$Y. Fujii is with Hamano Products Co., Ltd, Tokyo, Japan.}%
\thanks{$^{*}$\textit{These authors contributed equally to this work.}}
}

\begin{document}

\maketitle
\thispagestyle{empty}
\pagestyle{empty}

\begin{abstract}
This paper presents a feasibility study, including simulations and prototype tests, on the autonomous operation of a multi-limbed intra-vehicular robot (mobile manipulator), shortly MLIVR, designed to assist astronauts with logistical tasks on the International Space Station (ISS). Astronauts spend significant time on tasks such as preparation, close-out, and the collection and transportation of goods, reducing the time available for critical mission activities. Our study explores the potential for a mobile manipulator to support these operations, emphasizing the need for autonomous functionality to minimize crew and ground operator effort while enabling real-time task execution. We focused on the robot's transportation capabilities, simulating its motion planning in 3D space. The actual motion execution was tested with a prototype on a 2D table to mimic a microgravity environment. The results demonstrate the feasibility of performing these tasks with minimal human intervention, offering a promising solution to enhance operational efficiency on the ISS. 
\end{abstract}

\section{INTRODUCTION}
\label{sec:intro}
\subsection{Background}
\label{subsec:background}
Human-robot collaboration in space activities is expected to enhance the capabilities and efficiency of space exploration and utilization missions. While unmanned exploration applies robotic remote-control technologies, human spaceflight operations still have the potential to improve efficiency and capabilities through enhanced human-robot partnerships. 
The ISS has been instrumental in scientific and engineering advancements, with astronauts working on board. Robotic assistance is used to reduce the risk of extra-vehicular activities (EVA) with robotic arms. Yet, intra-vehicular activities (IVA) are heavily dependent on astronauts. Humans on board spacecraft can flexibly conduct various tasks with high dexterity, allowing continuous improvement of experiments and maintenance operations. However, the human work resource in space limits the possibilities for such operations~\cite{ISS_Crew-time}. Automating repetitive tasks through intelligent robotic systems will optimize and expand the possibilities of human spaceflight operations.

\begin{figure}[t]
\centering
    \includegraphics[width=\linewidth]{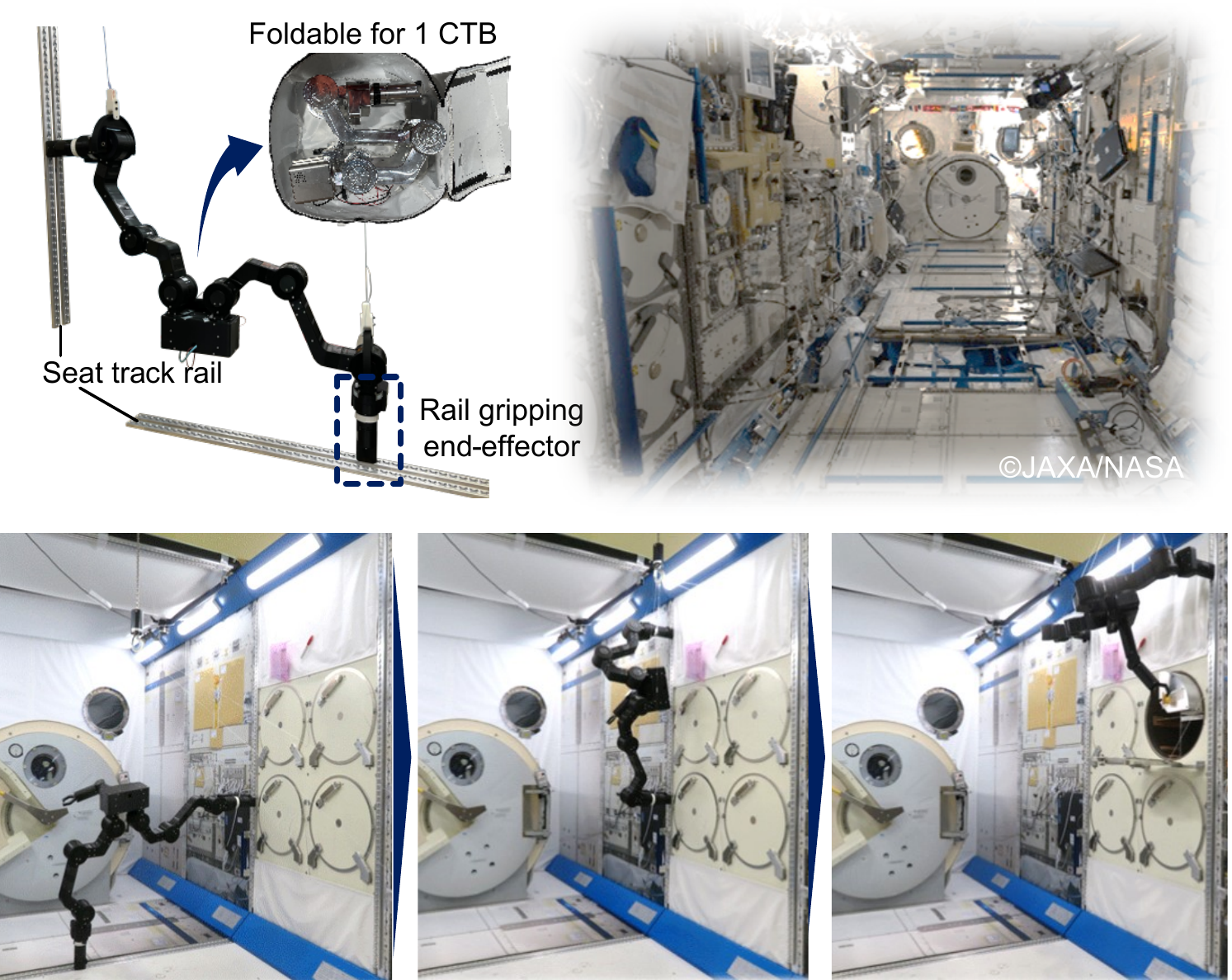}
    \caption{The Multi-Limbed Intra-Vehicular Robot (MLIVR) dedicated to assisting astronauts with routine tasks, such as cargo handling and task preparations. The robot moves by grappling ISS pre-existing seat-track interfaces.}
    \label{fig:concept}
\end{figure}

Current operational robots applied in human spaceflight mostly rely on remote control from the ground. For instance, the Space Station Remote Manipulator System (SSRMS) and the European Robotic Arm (ERA) can handle payloads outside the space station, with the ability to ``inch-worm walking'' by attaching themselves to dedicated grapple fixtures. 
These systems are primarily teleoperated, with human operators on the ground planning and executing movements. Given the communication delays and real-time human-robot collaboration needs, the autonomous operation of space robots is highly desirable. 
\begin{figure*}[t]
  \centering
  \includegraphics[width=\linewidth]{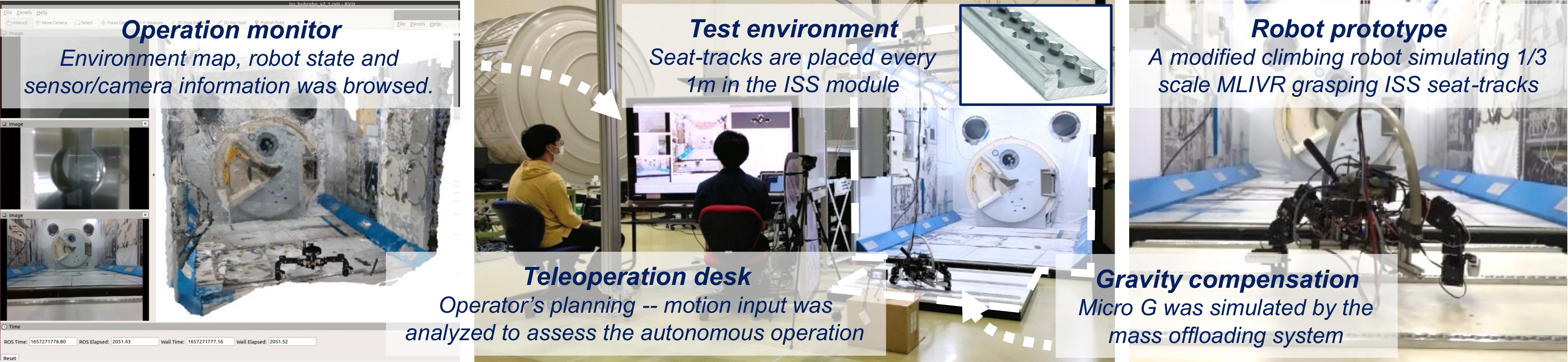}
  \caption{Concept of operation: preliminary testing to demonstrate a real operation from the ground to the space station \copyright JAXA/Tohoku Univ. Space Robotics Lab.}
  \label{fig:initial_tests}
\end{figure*}

\subsection{Related works}
\begin{table*}[b]
\centering
\caption{Comparison of the different types of the mobile EVRs and IVRs deployed in ISS.}
\label{tab:pastIVRs}
\begin{tabular}{l|llll}
\hline
                                     & \multicolumn{1}{c}{Free Flyers}     &  \multicolumn{1}{c}{Mobile Manipulators} & \multicolumn{1}{c}{Humanoids} \\
\hline
EVR Example                   & ~~AERcam~\cite{aercam}            & SSRMS~\cite{ssrms}, ERA  & -          \\
IVR Example                   & \begin{tabular}{l}Int-Ball~\cite{Int-Ball}, Astrobee~\cite{Astrobee},\\ CIMON~\cite{CIMON}\end{tabular}                   & - & Robonaut~\cite{R2}, SkyBot   \\
Advantages               & ~~Fast and agile mobility                 & Integrated mobility and manipulation          & Human I/F commonality                    \\
Limitations             & ~~Limited manipulation capability         & Required fixation point      & Volume-occupying \\
\hline
\end{tabular}
\end{table*}
Internal free-flyers, such as NASA's Astrobee~\cite{Astrobee} and JAXA's Int-Ball~\cite{Int-Ball, Int-Ball2} are capable of translating within the ISS module and collecting visual and sensory data. However, to provide more comprehensive support of astronauts, the integration of manipulation capability is desirable. While free-flyers have an advantage in their rapid translation capabilities, they are limited in terms of manipulation, especially when high power is required (see \tab{tab:pastIVRs}). Past technical demonstration projects on the ISS with humanoid robots, such as NASA's Robonaut2 (R2) ~\cite{R2} and ROSCOSMOS's SkyBot F-850, aimed to demonstrate such use cases. Yet, these robots have not yet been integrated into actual operations. Robonaut2 was developed to demonstrate astronaut support capabilities. It was initially deployed with an upper-body capable of dexterous manipulation and was tested on the ISS. Self-transportation capability was planned to be added to the lower-body. Two ``legs'', dedicated to the translation by grasping handrails inside the ISS, were developed ~\cite{R2_mobility}. Its ground tests validated the transnational capabilities, focusing on its stability and grasping. In terms of its autonomy, based on the constraint conditions and desired fixture points inputs from the user, the motion of each limb was planned and handrail rendezvous was realized~\cite{R2_autonomy}. 
Mobile manipulators, combining translation and manipulation capabilities have been used in the space station extra-vehicular environment. While existing mobile manipulators, such as SSRMS or ERA, are large inch-worming robotic arms, intra-vehicular robot is required to be small, non-invade size with more dexterous capabilities, and preferably share the existing ISS interior interface. Current ISS EVRs also require dedicated grapple fixtures for translation and grasping payloads. When applied to the intra-vehicular environment this could become a limitation in space and payload design. Thus, utilization of the existing crew interface - such as handrails or seat-tracks on the racks is anticipated.
Furthermore, due to the limited communication and ground operation cost, the autonomous operation is a key functionality for effective operations~\cite{ResileincySpace}. Multi-limbed robots, in ground application researches, advances in integrated path planning, foothold planning, and gait planning~\cite{HubRoboGaitPlanning}. Applying such methodology can improve autonomous operation minimizing ground control's planning and improving its operability and task execution. 

\subsection{Objectives and Contributions}
\label{subsec:objectives}
This paper proposes an autonomous operation for the Multi-Limbed Intra-Vehicular Robot (MLIVR), focusing on its translation capabilities, integrating autonomous foothold and gait planning until its execution. The main objective is to evaluate the feasibility of such autonomous translation for the MLIVR within the ISS through a combination of simulations and ground tests.


\section{APPROACH}
\label{sec:method}
\subsection{Desired Operation}
Desired tasks for IVR assistance were analyzed based on the current human-in-the-loop operation of space stations~\cite{JEMtaskAnalysis}. Astronauts need to prepare and clean before/after most tasks. The equipment is normally kept in standard transfer bags, called \textit{cargo transfer bag} (CTB), which need to be collected and transported where the operation takes place. ISS being an on-orbit laboratory for science utilization (experiments), significant crew time is spent on configuration changes and payload swaps.
To realize such tasks, the robotic system must efficiently travel within the space station, including manipulation capabilities, and gather data for transmission to the ground. IVRs are anticipated to use existing interfaces, such as seat-tracks on ISS racks, for potential anchoring points. Such concept of MLIVR was considered and studied in the previous work \cite{PORTRS24}. One of the key issues to applying such a robot in the actual operation is its level of autonomy to minimize the crew work, operator effort and ease operations. 

\begin{figure*}[t]
  \centering
  \includegraphics[width=\linewidth]{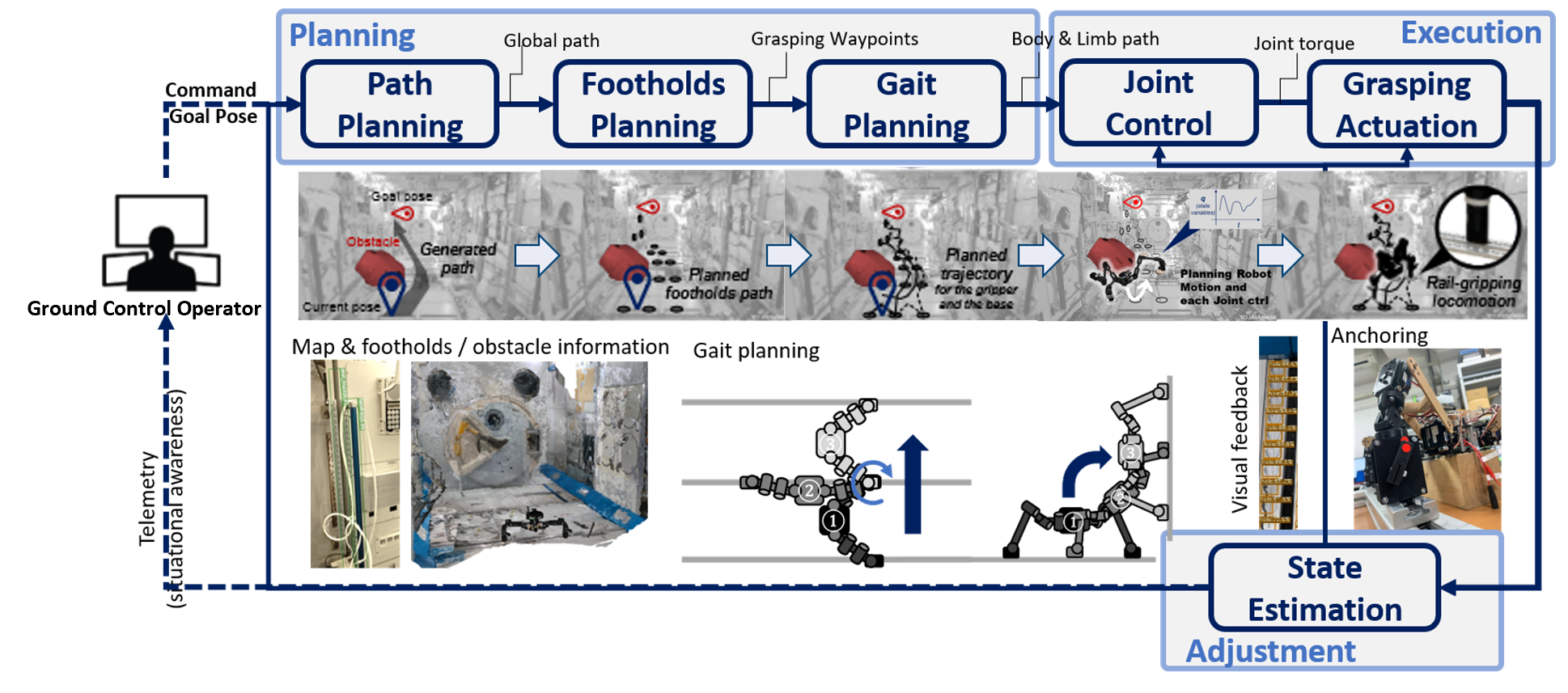}
  \caption{Concept of the autonomous translation of the multi-limbed intra-vehicular robot operation.}\label{fig:ops_flow}
\end{figure*}

Initially, 1/3 scale prototype was developed based on a previously developed legged climbing robot \cite{Uno2021_HubRobo}, and tested under remote control with manual planning in the analogue test cite (see \fig{fig:initial_tests}). It was envisioned that the existing ISS interface - seat tracks- would be used for its fixation. With the ground tests, it was realized that the teleportation of the end-effector to adjust and fixate to the seat-track, required significant time and effort for the operator.  Thus, we developed a new gripper that can better compensate for position error and defined the following concept of operations (ConOps) for the autonomous rail-gripping mobile manipulator in the space station (also see \fig{fig:ops_flow}).
\begin{itemize}
    \item \textit{Planning}: Robot system autonomously computes path, foothold, and gait based on the known parameters and environment data. 
    \item \textit{Execution}: Robot performs limb movements, grasps footholds, and adjusts grasping based on sensory feedback.
    \item \textit{Adjustments}: While translating and fixating it gathers environment information, and if the planned path is found to be obstacles it re-plans the path. Repeat the above steps until the desired goal is achieved.  
\end{itemize}

Given the desired goal input, the robot system plans its global path to navigate from its current position to the destination. Using the robot's location and environment information, its translation path is planned. Available grasping points (footholds) are considered to plan each sequence motion of grasping and moving to the next footholds (foothold planning). Next, the robot plans a gait pattern and posture that determines the sequence of which leg to use, the posture for grasping, and how to move while grasping the planned grasping points. Based on this plan, the robot executes the motion trajectories of the legs and torso. Once the robot's foot reaches the grasping point, it executes the action to anchor itself to the foothold. The success of the grasping can be confirmed using sensory information (visual and tactile). The environment information and the robot's pose are updated based on the sensory information, such as joint encoders, and visual and inertial measurement units (IMUs). Based on the updated self-localization, the path to the destination is re-planned as needed (e.g., when new obstacles are recognized on its planned path), and repeated until the goal is reached. This operation flow is visually represented in the \fig{fig:ops_flow}.

\subsection{Evaluation method}
To evaluate its feasibility, a combination of simulation and ground prototype testing was applied. The planning sequence and its algorithm feasibility for path planning and foothold planning need to be considered in 3D space. To simulate such motions in a microgravity 3D environment, ClimbLab~\cite{uno2021climblab}, MATLAB-based legged climbing robot simulator, was employed. The simulation computes the desired path with each foothold and torso's desired poses and outputs each joint torque command for a desired motion. These motions are validated with a prototype model described in the following chapter. The developed prototype will be tested with ground testing validating each joint motion realizing translation within the ISS module integrated with grasping motion.  



\section{MULTI-LIMBED INTRA VEHICULAR ROBOT}


The design of the prototype incorporates concepts for autonomous translation while considering launch and safety constraints. The prototype is equipped with structures and mechanisms to ensure harness safety, accommodate sudden external loads, and maintain emergency crew evacuation routes. The primary design requirements include:

\begin{itemize}
    \item Launch mass and volume: The robot's total mass is targeted to be less than 15 kg, with a stowage volume that allows it to be folded into a single cargo transfer bag (CTB).
    \item Grasping translation interface: The robot is equipped with a gripping mechanism capable of grasping existing onboard fixation points, such as seat tracks or handrails on ISS standardized racks 
    \item Translation method: The robot should be capable of striding along rails laid parallel on a surface and transitioning to rails on adjacent vertical surfaces. Seat-track rails are 2\;m long, placed every 1\;m.
\end{itemize}

\subsection{Hardware prototype}
The MLIVR prototype was developed to test translational capabilities. While the actual system is envisioned to perform both translation and dual-arm manipulation, the prototype focuses solely on testing translational movement in a simulated two-dimensional microgravity environment. When moving along a plane, while keeping the base parallel to the bottom, the necessary joint axis configuration for moving within the same plane by grasping rails requires 5 degrees of freedom per leg. To achieve a compact and lightweight design, the prototype was designed with two legs, sufficient for its translation needs. Additional limbs can be added if required for manipulation capability or redundancy.

\begin{figure}[t]
  \centering
  \includegraphics[width=\linewidth]{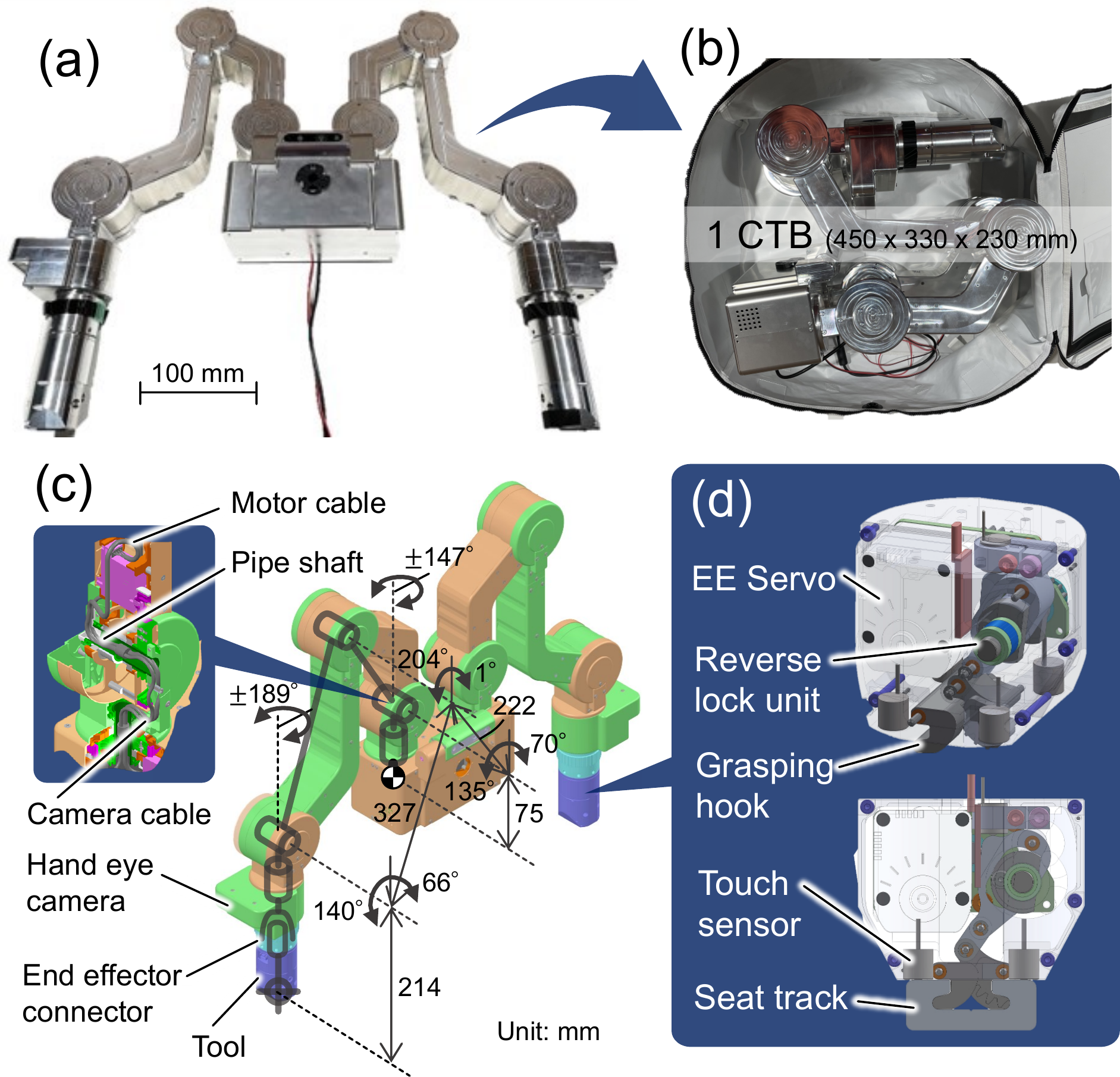}
  \caption{Prototype of the intra-vehicular robot (top) and the concept of the rail-grasping locomotion and cargo handling within the interior of the space station. The robot is designed to be foldable, allowing it to fit into a soft bag for shipment and minimize space impact in the ISS.}\label{fig:biped_ivr_mechanical_structure}
\end{figure}

The prototype is designed to grasp the seat tracks using a developed gripping mechanism. The gripping mechanism, presented in the \fig{fig:biped_ivr_mechanical_structure}(d), fixates the robot to the seat-track by inserting the grasping hook into the rail groove. This gripping mechanism is designed to compensate for positioning errors of the end-effector. It can fixate to any point along the seat-track length and compensate for more than 15 degrees of error in its yaw rotations. 

\fig{fig:biped_ivr_mechanical_structure}(a) illustrates the nominal posture of MLIVR during its activities. 
When fully extended, the stride of the robot is 1200 mm within the same plane. When folded, the robot can be compressed to 450 mm x 330 mm x 230 mm (see \fig{fig:biped_ivr_mechanical_structure}(b)). The materials used were chosen to prevent potential fire scattering or off-gassing inside the ISS. 
The exterior and frame of the robot are made almost entirely of aluminum alloy. By enclosing all mechanical and electrical components within a metal exterior, the design takes into account fire prevention and the prevention of component scattering in the event of damage within the ISS. 
The total weight of the prototype is 9.2 kg. When additional devices for autonomous operation in orbit are included, the total weight is expected to be 10.6 kg. Both the expected weight and the folded size ensure that the developed intra-vehicular mobility robot meets the requirements for packaging and transportation in a CTB, similar to ISS consumables.

\fig{fig:biped_ivr_mechanical_structure}(c) shows the structure of the robot's limbs. The robot's legs have five joints, with the units shown in green and orange driven independently. 
The harness path at the robot's joint is shown at the upper left of Figure \fig{fig:biped_ivr_mechanical_structure}(c). 
To minimize bending of the harness stored inside the legs and reduce stress during joint movement, each joint shaft is hollow, allowing the harness to pass through it. Additionally, to prevent wear debris generated from the sliding contact between the cables and the robot frame, highly lubricative resin covers are provided at the sliding points. 

The end-effector, indicated in blue in \fig{fig:biped_ivr_mechanical_structure}(c), is attached to the leg tip and functions to grasp the seat tracks. The detailed internal structure of this mechanism is shown in \fig{fig:biped_ivr_mechanical_structure}(d). This mechanism includes two claws that are operated by a servo motor, designed to prevent reverse input when the claws are fully extended. When the claws are deployed, they can lock the groove of the seat track from the inside, enabling a firm grasp of the seat track. Contact sensors are placed at the corners of the gripper base to detect the contact state between the gripper mechanism and seat track.
Additionally, hand-eye cameras are used to recognize the surrounding environment and provide feedback on the position of the leg tips, which is used for positioning through small movements of the leg tips (visual servoing). The microcontroller for actuator control drives each joint according to the target angles received from the onboard PC. It controls the joint servos and the actuators of the leg-tip tools through a communication conversion board and a communication and power hub board connected below it. All actuators of the robot are connected to the microcontroller via a single bus communication system. Through this bus communication system, the microcontroller can obtain temperature, estimated torque, and current angle from each joint servo, as well as the contact status with the seat track from the leg-tip tools, and transmit this information to the onboard PC.



\subsection{Software system}
The software architecture is based on the system used for the quadrupedal space exploration robot HubRobo \cite{Uno2021_HubRobo}, with three key modules: a high-level controller, a low-level controller, and a state estimator.

One of the critical requirements for the ISS intra-vehicular mobility robot is the ability to accurately position its gripper-equipped hand at the correct rail-gripping position during movement. To achieve this, an automatic adjustment function based on visual sensor feedback is implemented. This study employs an image-based visual servoing method \cite{visualServo} for the automatic adjustment process. The control law is shown in Equation \eq{eq:visual_servoing_control_law}.
\begin{equation}
    {(v_x, v_y, v_z, \omega_x, \omega_y, \omega_z)}^\top = - \lambda L^{\top +} (\bm{s}-\bm{s}^*) 
    \label{eq:visual_servoing_control_law}
\end{equation}
The left-hand side represents the translational and angular velocities of the camera in the camera coordinate system. The right-hand side includes $\lambda$, a constant gain, $L^{\top +}$, the pseudoinverse of the image Jacobian, $\bm{s}$, the feature vector(i.e., point), and $\bm{s}^*$, the target feature vector.


\section{Validation}
\subsection{Planner - Simulations}
\begin{figure}[t]
  \centering
  \includegraphics[width=\linewidth]{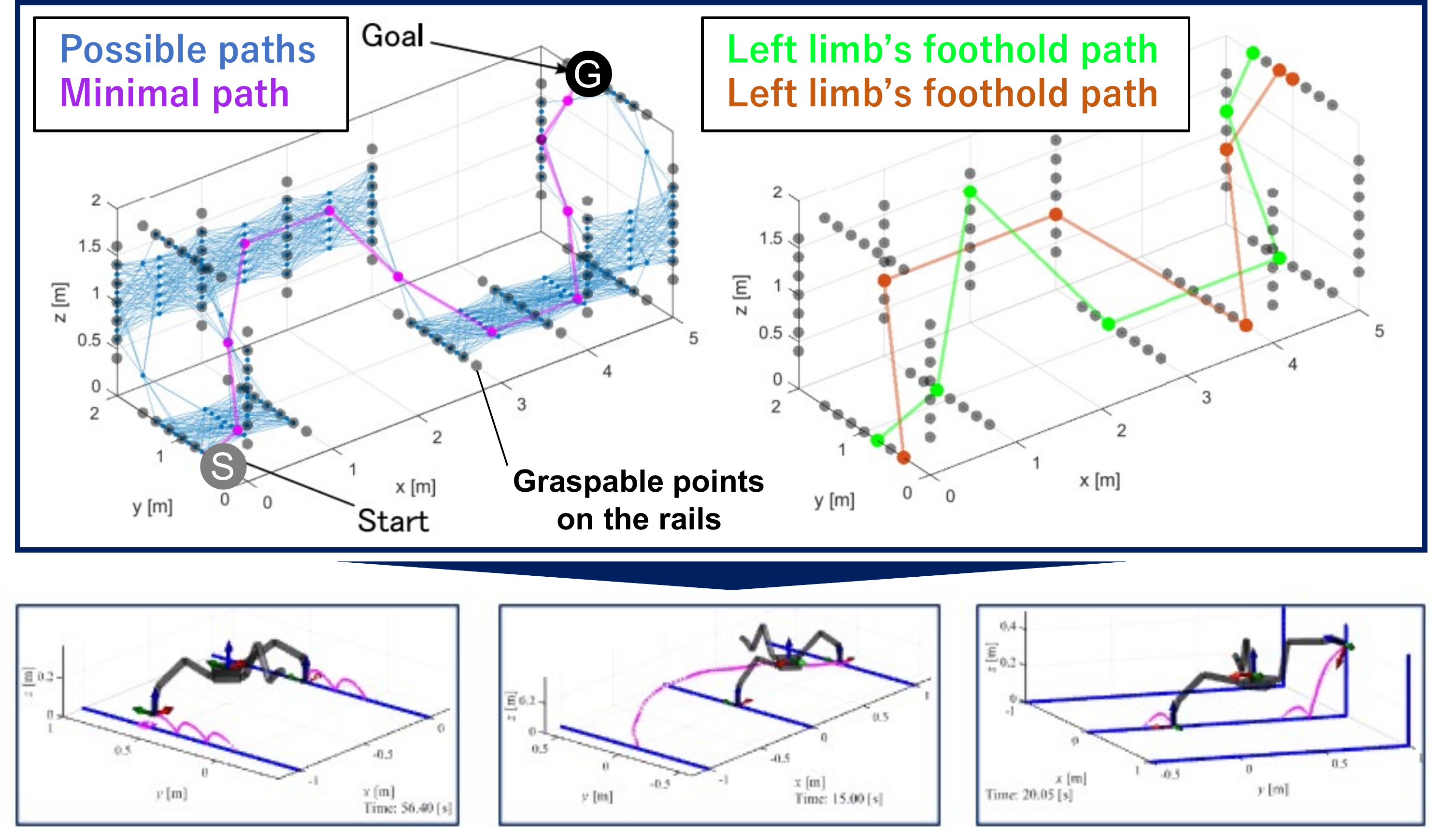}
  \caption{Simulated path and foothold planner (top) and the snapshots of the dynamic simulation of the MLIVR in our simulation platform: ClimbLab\cite{uno2021climblab} (bottom). Graph-based planner solves the shortest feasible path of the robot body (pink path) and the foothold of each limb (green and orange path). Based on the simulation joint motion plan (joint angles or torque) can be generated.}\label{fig:planner_sim}
\end{figure}

The validation of path planning, foothold planning, and gait planning was conducted with a simulation environment. In this setup, the locations of graspable seat-tracks inside the module and obstacles (non-graspable points) were predefined. Details of the simulation were presented in the previous study \cite{Imai_gait_ISS}. When the robot's grippable footholds are limited, which is a typical problem for legged climbing robots, path, foothold, and gait planning greatly interfere with each other because the kinematically feasible robot postures are severely constrained. Thus, some previous works considered such a planner for legged climbing robots to maintain the criteria such as tumble stability and kinematic reachability ~\cite{HubRoboGaitPlanning,HubRoboGaitPlanning2}. A graph-based path and foothold planner were proposed to compute the fastest path for the robot's base and footholds simultaneously \cite{takada2023graph}. \fig{fig:planner_sim} presents the generated path considering the foothold of the robot based on the initial and goal position inputs. Based on the foothold information, each joint trajectory was generated that could realize the computed gait. These generated joint motions were then utilized for the prototype's motion execution validation, as described in the following subsection.

\begin{figure}[t]
  \centering
  \includegraphics[width=\linewidth]{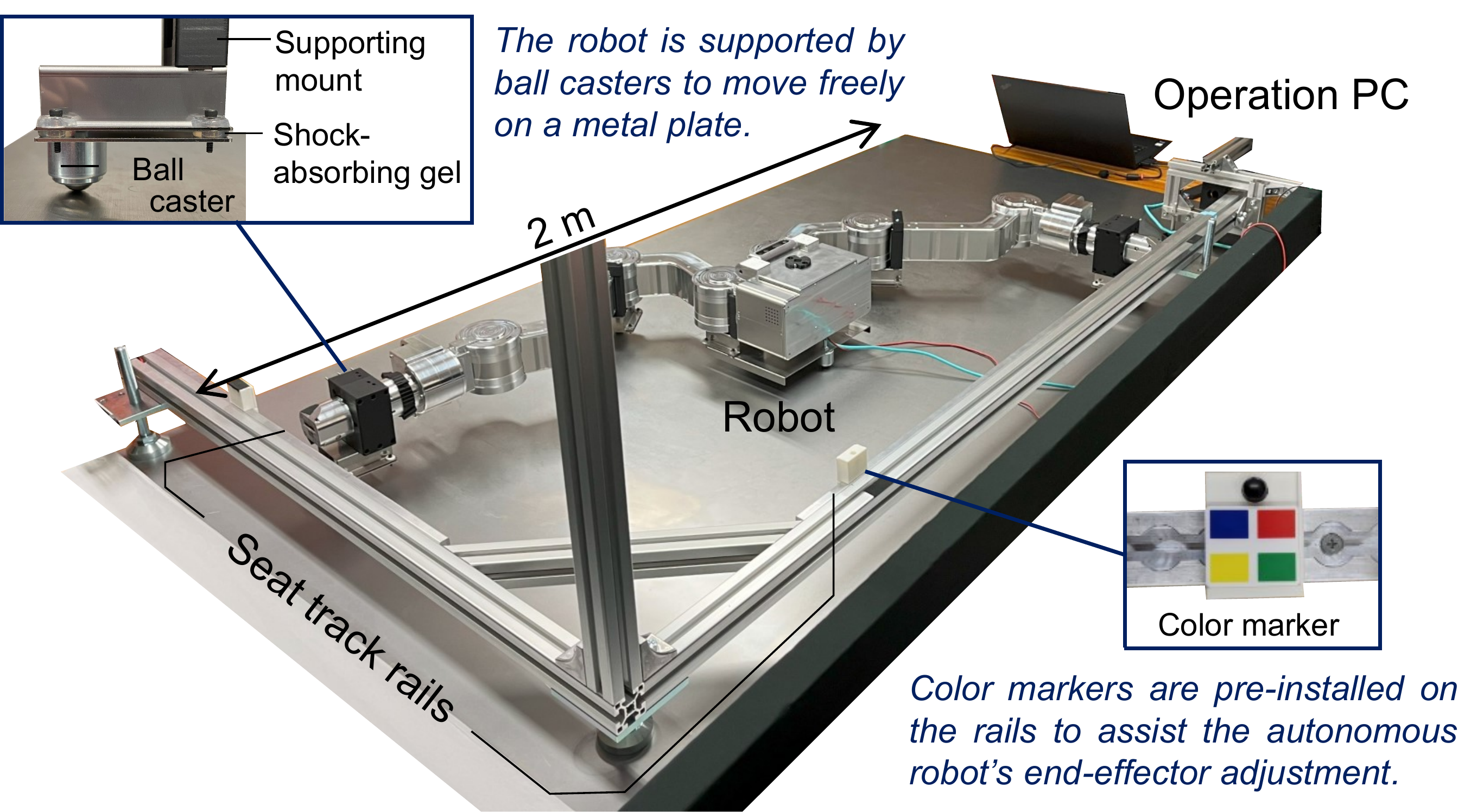}
  \caption{Two-dimensionally microgravity emulated test setup. The robot is supported through ball casters to neglect friction on the metal plate.}\label{fig:experiment_setup}
\end{figure}
\begin{figure*}[t]
  \centering
  \includegraphics[width=\linewidth]{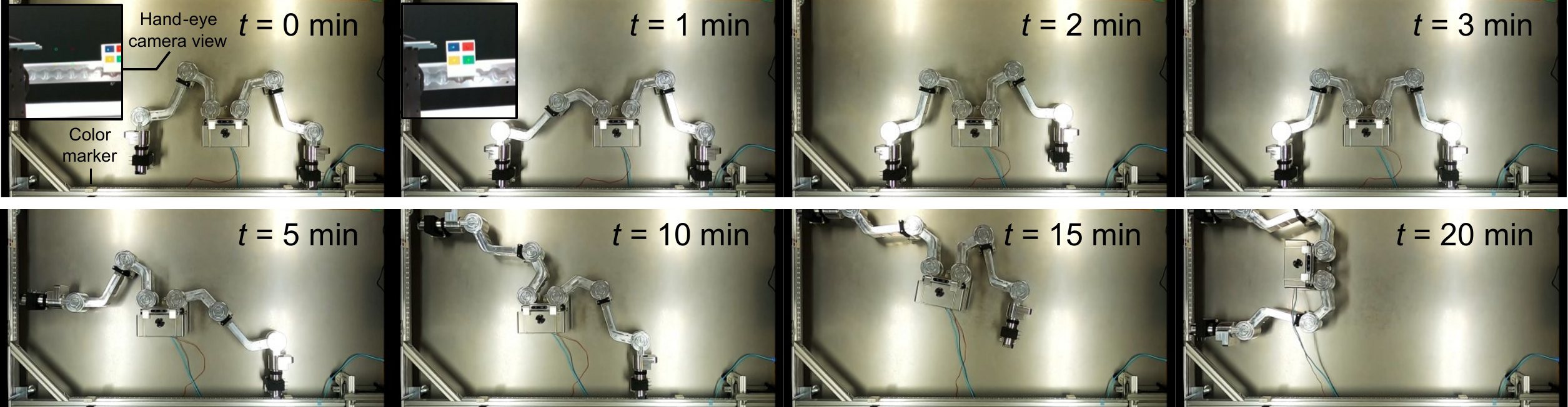}
  \caption{Capture of the autonomous multi-limbed intra-vehicular robot operation to travel on the vertically arranged rails on the two-dimensionally microgravity simulated test set up. Horizontal translation at $t=0~5$~min was executed autonomously with visual servoging, and vertical translation after $t=5$~min was executed with teleoperation. }\label{fig:experiment_result}
\end{figure*}

\subsection{Motion Execution - Ground Testing}
To verify the feasibility of the automated movements described in the previous section, ground validation experiments were conducted using a full-scale prototype. \fig{fig:experiment_setup} shows the testing environment prepared for this study. The robot on the metal plate is supported at multiple points by ball casters through an impact-absorbing gel. Because the friction between the ball casters and the metal plate is very low, the robot can move freely on the plate. Although the microgravity simulation accuracy of this experimental setup is inferior compared to air flotation on a granite surface plate, it is very simple, low-cost to develop and use, and suitable for long-duration experiments since it does not require compressed air. Additionally, color markers were used as the image features, extracting centroids of each color rectangle, for the visual servoing.
The sequence of the autonomous movement was defined as follows: 1) translational movement of the swing leg along the rail, 2) automatic adjustment of the hand's position and orientation using visual servoing, 3) application of a preset force in the normal direction of the rail, and 4) gripping. By executing these sequences in order, autonomous stride movement was attempted. In this experiment, the amount of swing leg translational movement in step 1) was predetermined, considering the reach of one leg of the robot, and the target leg tip position based on the base coordinate system was commanded step by step for the experiment. For safety, the execution commands for each sequence were operated after confirming the safety of the robot and equipment (which resulted in the long experiment time compared to the actual motion time of the robot).

\fig{fig:experiment_result} shows the ground validation experiment of the autonomous intra-vehicular mobility robot. The state of the color markers as seen from the hand-eye camera during the automatic control of the hand position using visual servoing is shown at times $t=0$~min and 1 min. The visual servo converged at $t=1$~min. In this experiment, the robot performed an autonomous stride by sequentially executing the sequence until the convergence of the visual servo for the third step, where the leg extends to the vertical wall ($t=5$~min). t is important to note that, on average, it took 4 minutes to complete one step (execution and adjustment) of the robot during the preliminary analog teleoperation testing (Fig. \ref{fig:initial_tests}), indicating that this sensory autonomous execution was four times more efficient. Starting at $t=5$~min, the operation was switched to manual control, where vertical rail translation was tested. The robot successfully transitioned to the vertical rail ($t=20$~min), indicating the feasibility of the movement and requiring a similar execution time for 1 step, around 5 minutes each. Applying the same automation methodology, it is anticipated that the execution time could be significantly shortened.   Throughout this ground validation experiment using the full-scale prototype of the developed ISS intra-vehicular mobility robot, we confirmed the appropriateness of the control flow constructed for autonomous and automatic movement along the rail in a microgravity environment and the feasibility of this robot system.

\section{CONCLUSIONS AND FUTURE DIRECTIONS}
This research explored the autonomous transportation of the multi-limbed intra-vehicular robot (MLIVR), analyzing and validating task motions and operational sequences through simulations and ground-prototype testing. The study proposed and demonstrated a computational flow for autonomous operation. Starting with an operator-defined goal pose, the system successfully planned paths, foothold positions, and gaits in a simulation environment. These results were then validated with a physical prototype tested on a 2D emulation platform. The validation included autonomous limb manipulation, body movement, and foothold grasping, utilizing visual feedback from a hand-eye camera. The integrated process demonstrates the feasibility of deploying such a robot for autonomous operations on the ISS.

The overall integrated process presents the feasibility of deploying such a robot in autonomous operation. Using a developed prototype and autonomous translation algorithm, we plan to further evaluate quantitative metrics of the operation time, precision of movements, and energy efficiency in comparison with remote control teleportation. It will be also required to asses failure modes during the operation to apply autonomous detection and recovery.  Furthermore, integrating existing robotic solutions, such as obstacle recognition within SLAM (Simultaneous Localization and Mapping) or target object manipulation, would enable such robots to transport goods and prepare astronauts' tasks on board the ISS. At the same time, they rest or conduct other tasks. The proposed robot configuration considered a launch mass, volume, power, and other constraints within the ISS. Demonstrating such a system on the ISS will gather valuable data for future robotics applications to serve in the actual crew task assistance. Such a system is proposed as JAXA's Payload ORganization and Transportation Robotic System (PORTRS)\cite{PORTRS24}. To minimize the ground operation cost, PORTRS should be able to execute pre-defined movements such as translation autonomously as tested in this research. While the joint configuration required to perform a given manipulation task might be updated, the autonomous operation flow proposed should be common. Furthermore, it will integrate the manipulation capability of equipment swap to reduce the crew workload in preparation and closeout tasks. Moreover, a mobile robot working on its batteries should be able to perform onboard power management and capability to dock and re-charge by itself. In this process, the same translation method could be applied. When applying such robots in further exploration applications, such as the Lunar Gateway\cite{gateway} or within the transportation spacecraft, human presence will be further limited. One of the prospects for the Gateway is to include IVR for its maintenance. Within its development, the translation path and space for fiducial markers are considered. The results of this study present that with these assumptions, MLIVR should be able to operate within the unmanned period of the Gateway. In such a case, high reliability of the system will be required. ISS tests of such systems will contribute to asses such metrics in the real application environment.

To further enhance the usability and efficiency of these robots, several areas for improvement have been identified. First, task execution speed could be improved by optimizing robot motion speed, computation time, and decision-making pace. During our tests, verifying the correctness of each movement was most time-consuming. Self-verification and adjustment of the movement and grasping are crucial to execute autonomous operations. Another approach of the compliant control to mitigate the inertial force while cargo manipulation \cite{imai2024CLAWAR}, which is a dominant impact on the robot in microgravity, is also considered. It is important to mention that vision-based direct recognition of the rails is also feasible based on our preliminary trial while the color marker was utilized in the last testing in this paper. Second, size minimization is anticipated if such robots should work side by side with astronauts. The MLIVR's current sizing was set to enable a 1-meter stride in movement. With the currently proposed method, the addition of rails (such as crew-used handrails) perpendicular to the rack could help minimize robot size. Another approach might be to enable the robot to translate without anchoring its base at all times - controlling its translation as a free-flier, which would require further research to operate safely in microgravity. For planning, employing an external computational device on board the ISS could help minimize the size and power consumption of the robot. Finally, further application development in terms of manipulation task execution is anticipated, including dexterous task execution as well as human collaboration. The number of joints is a trade of mass/volume/power with dexterity and ease of control. While the prototype tested had 5DoF on each limb, considering ease of control in the ISS environment where many cables and obstacles are present, 6-7 DoF from the body could ease the manipulation application.  The deployment of such robots in space stations could also benefit further development and testing of these technologies for future human spaceflight activities.


\section*{Acknowledgment}
This work was conducted as a joint research project between Tohoku University and JAXA, with partial support from JSPS KAKENHI Grant Number JP23K13281. The authors would like to express their gratitude to Mr. Mikio Eguchi, Mr. Koki Murase, Mr. Takuya Kato, and the SRL-Limb team for their significant support in the experiments and software development. Special thanks to Mr. Riichi Itakura for his contributions to joint research efforts and project management. We also extend our appreciation to Mr. Masaru Wada, Mr. Tetsuya Inagaki, and Dr. Akiko Otsuka from JAXA for their expertise in space robotics applications. We thank Mr. Hiroya Uchida, Mr. Fumiya Miyachi, and Hamano Products Co., Ltd. for their invaluable assistance in the robot hardware design and prototyping.

\bibliography{./IEEEabrv,bibliography.bib}

\end{document}